\documentclass[letterpaper, 10 pt, conference]{ieeeconf}
\IEEEoverridecommandlockouts
\usepackage{graphicx}
\usepackage{caption}
\usepackage{subcaption}
\usepackage{xcolor}
\usepackage{multirow}
\usepackage{tabularx}
\usepackage{color, colortbl, soul}
\usepackage{multicol}
\usepackage{listings} 
\usepackage{tcolorbox}
\let\labelindent\relax
\usepackage{enumitem}
\usepackage{booktabs}
\usepackage{siunitx}
\usepackage{amsfonts} 
\usepackage{cite}
\usepackage{hyperref}
\hypersetup{
    colorlinks=true,  
    citecolor=red,    
    linkcolor=red,    
    filecolor=magenta, 
    urlcolor=blue     
}

\definecolor{lightgreen}{rgb}{0.8,1,0.8}

\title{\LARGE \bf \textsc{InteRACT}: Transformer Models for Human Intent Prediction \\ Conditioned on Robot Actions }

\author{Kushal Kedia$^{1}$, Atiksh Bhardwaj$^{1}$, Prithwish Dan$^{1}$, Sanjiban Choudhury$^{1}$
\thanks{$^{1}$Department of Computer Science, Cornell University}}%

\begin{document}

\maketitle
\thispagestyle{empty}
\pagestyle{empty}

\begin{abstract}
In collaborative human-robot manipulation, a robot must predict human intents and adapt its actions accordingly to smoothly execute tasks. However, the human's intent in turn depends on actions the robot takes, creating a chicken-or-egg problem. Prior methods ignore such inter-dependency and instead train \emph{marginal} intent prediction models independent of robot actions.
This is because training \emph{conditional} models is hard given a lack of paired human-robot interaction datasets. 

Can we instead leverage large-scale human-human interaction data that is more easily accessible? Our key insight is to exploit a correspondence between human and robot actions that enables transfer learning from human-human to human-robot data. We propose a novel architecture, \textsc{InteRACT}, that pre-trains a conditional intent prediction model on large human-human datasets and fine-tunes on a small human-robot dataset. 
We evaluate on a set of real-world collaborative human-robot manipulation tasks and show that our conditional model improves over various marginal baselines. We also introduce new techniques to tele-operate a 7-DoF robot arm and collect a diverse range of human-robot collaborative manipulation data which we open-source. We release our code and datasets at \url{https://portal-cornell.github.io/interact/}.
\end{abstract}

\section{Introduction}

\begin{figure}[t!]
    \centering
    \includegraphics[width=8.5cm]{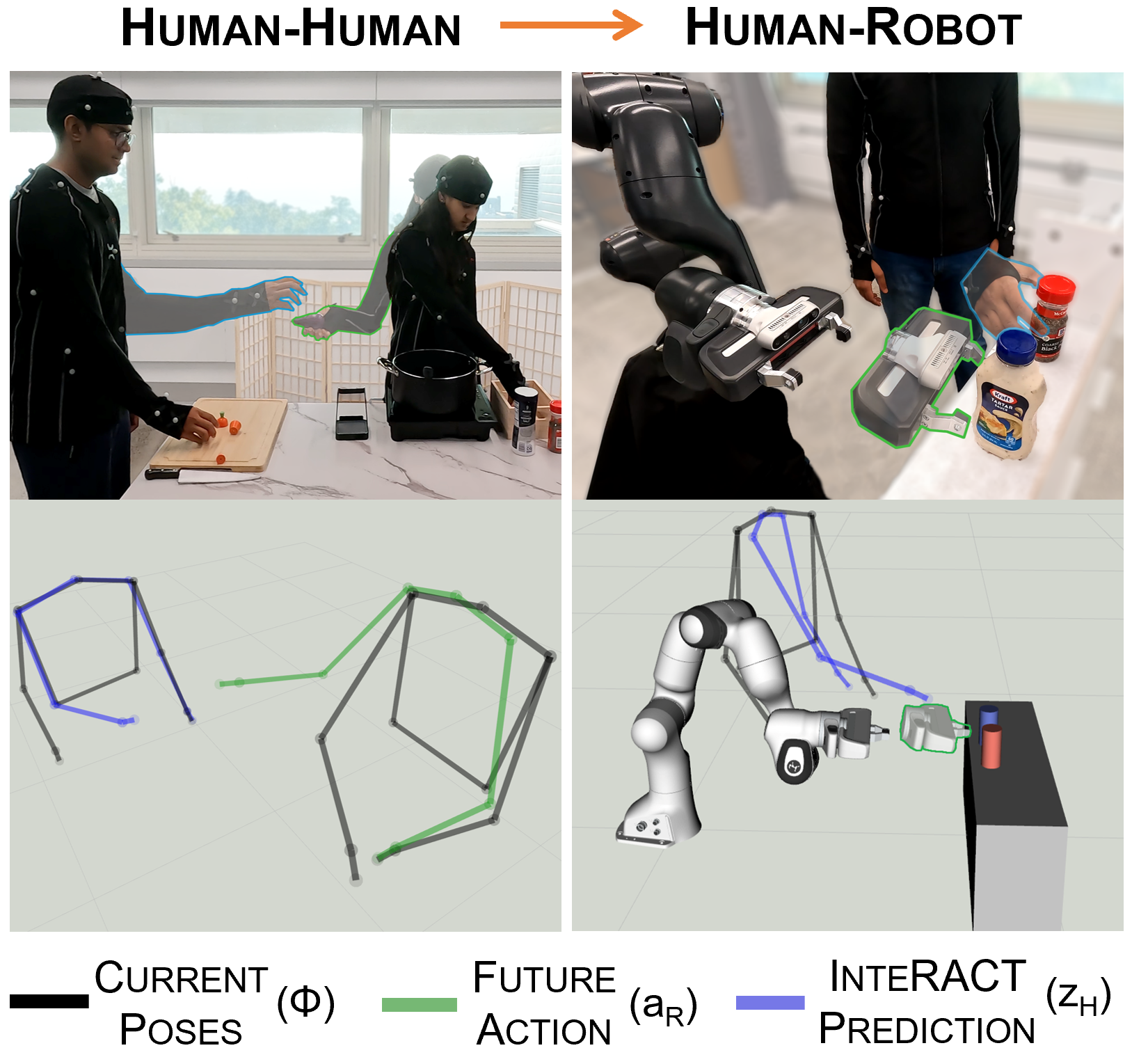}
    \caption{We present \textsc{InteRACT}, a model that predicts future human intent conditioned on the future robot action. \textbf{Left:} When a human passes an object over,  \textsc{InteRACT} conditions on the future object handover action of one human and predicts that the other human will move towards it. \textbf{Right:} In this human-robot interaction, given the robot's plan to reach for the can on the right, \textsc{InteRACT} predicts the human will reach for the pepper. We transfer a model trained on human-human interactions to human-robot interactions.}
    \label{fig:fig1}
    \vspace{-7mm}
\end{figure}

If robots are to work alongside human partners to achieve shared goals, they need models for how to coordinate with humans. Such coordination is dependent on understanding the human partner’s intent and predicting how these intents might change in response to the robot’s actions~\cite{dragan2017robot}. Consider the shared human-robot manipulation task in Fig.~\ref{fig:fig1} where a human and a robot are simultaneously reaching for objects on a shelf. The robot needs to predict the human's intent, i.e., which object they are reaching for, to safely and confidently reach for a different object. However, the human's intent in turn depends on the action the robot takes in the future. This cyclic dependency between human intent and robot actions presents a non-trivial chicken-or-egg problem. We tackle the problem in this paper by training intent prediction models that condition on future robot actions. 

There's been a lot of recent focus on intent prediction for collaborative manipulation~\cite{Gui2018TeachingRT,laplaza2021attention,zhang2020recurrent,laplaza2022contextattention}, including approaches~\cite{kedia2023manicast} that leverage large-scale human-activity datasets~\cite{AMASS:ICCV:2019, ionescu2013human3}. Nevertheless, these models predominantly operate in a \emph{marginal} framework, without conditioning on future robot actions. Such an approach can yield sub-optimal outcomes; consider again the scenario illustrated in Fig.~\ref{fig:fig1}. An unconditioned model may estimate that the human has an equal likelihood of reaching for either object on the shelf. Consequently, the robot may deduce that it is unsafe to proceed with reaching for any object.

Conditional transformer models show promise in overcoming such issues and have been successfully used in self-driving~\cite{Ngiam2022SceneTA, Huang2022ConditionalPB, Huang2023LearningIM, Song2020PiPPT, Tolstaya2021IdentifyingDI} to model dependencies between road agents and forecast their joint behaviors. Such models require extensive human-generated driving data~\cite{Ettinger2021LargeSI,Zhan2019INTERACTIONDA}. However, adapting such methods to the domain of human-robot collaborative manipulation is not straightforward due to a key obstacle: the scarcity of large-scale human-robot interaction datasets for training. Acquiring such datasets, even on a smaller scale, poses its own challenges, given the complexity of teleoperating 7-DoF robot arms. The question then arises: can we capitalize on the readily available, large-scale human-human interaction data?

\textbf{\emph{Our key insight lies in leveraging the correspondence between human and robot actions to facilitate transfer learning from human-human to human-robot interactions.}} For example, in common manipulation tasks such as object handovers, humans often discern each other's intentions by observing arm and hand movements. We hypothesize that human reactions to robot arm movements exhibit similar patterns, allowing for the effective transfer of learned models.

We propose a novel architecture, \textsc{\textbf{InteRACT}} (\textbf{Inte}nt Prediction via \textbf{R}obot \textbf{A}ction-\textbf{C}onditioned \textbf{T}ransformer) that can predict a human's intent based on the robot's planned future action. Our model is trained in two stages. First, we utilize large sources of both single and multi-human interaction data, where our model predicts human intent conditioned on the future action of the other human in the scene (Fig \ref{fig:fig1}). Then, we exploit a low-level correspondence between the human's hand and the robot end-effector to tele-operate a 7-DoF Franka Emika robot arm alongside a human partner. This collected Human-Robot dataset contains human-robot interaction data as well as the corresponding motion data of the human tele-operating the human arm. We utilize this pairing to align human and robot representations for effective transfer learning. Our key contributions are:
\begin{enumerate}[nosep, leftmargin=0.3in]
    \item We introduce a novel transformer-based architecture that conditions on robot actions to predict human intent.
    \item We propose a technique to collect a paired human-robot dataset via tele-operation for fine-tuning models with aligned representations and open-source a first dataset of human-robot collaborative manipulation.
    \item Our prediction model demonstrates improved human intention prediction on multiple real-world datasets of human-human and human-robot interaction.
\end{enumerate}

\section{Related Work}

{\bf Predicting Human Intent for Navigation.} 
Human intent prediction has been extensively studied in social navigation. Research in this domain has focused on developing better input and output representations for modeling inter-agent interactions~\cite{chen2019crowd, Chen2022SafeNav, monti2021dag}. The multi-modality of human intents can be captured by generative neural network architectures such as Trajectron++~\cite{salzmann2020trajectron++,mavrogiannis2022winding,poddar2023crowd, kothari2023safety}. Yet, these works are largely independent of robot actions and predict human intents independently for each agent in the scene without enforcing joint future consistency. Attempts have been made to develop joint forecasting and planning frameworks~\cite{Tian2021SafetyAF, Kedia2023AGF}. Our work is inspired by recent progress in the self-driving domain; transformer models have been applied to predict the joint futures of agents in the scene by conditioning the robot's future actions~\cite{Ngiam2022SceneTA, Huang2022ConditionalPB, Huang2023LearningIM, Song2020PiPPT, Tolstaya2021IdentifyingDI}. While such an approach is feasible in self-driving where large-scale interaction datasets~\cite{Ettinger2021LargeSI,Zhan2019INTERACTIONDA} are readily available, it is difficult to collect human-robot collaboration data. We introduce new techniques to collect a dataset of human-robot interactions to fine-tune forecasting models trained on human-human interactions.

{\bf Human Pose Prediction.} In this work, we represent human intention as a trajectory of human pose predictions, which is a challenging problem due to the wide range of possible human joint movements. Recently, the release of large-scale datasets of human motion~\cite{AMASS:ICCV:2019,ionescu2013human3}, has made this problem more tractable, leading to rapid progress in this field. A number of approaches have been proposed to model the spatio-temporal interaction between human joints using Graph Neural Networks and Transformers~\cite{Mao2020HistoryRI,Sofianos2021SpaceTimeSeparableGC}. However, such approaches are limited to predicting future motion for just one human. Recent works have extended pose prediction from single-person to multi-person settings. SoMoFormer~\cite{vendrow2022somoformer} uses a transformer architecture that can accept any joint embedding as a query, allowing the model to learn interactions between all joints in the scene, including from different humans. Multi-Range Transformers (MRT)~\cite{Wang2021MultiPerson3M} utilizes a combination of a local encoder to learn temporal dependencies between a single agent's body pose and a global encoder to learn dependencies with other agents in the scene. The small sizes of existing human-human interaction datasets limit these approaches. In this work, we extend existing human-human interaction datasets and additionally collect a dataset of human-robot collaboration.

{\bf Predicting Human Intent for Collaborative Manipulation.} 
Human-robot collaboration requires representing future human intents in some form. Many approaches consider the human to be completely static~\cite{yang2022model,sisbot2012human,lasota2014toward,liu2021collision}. Others reason about the future motion of specific human joints such as the wrist, hand, or head~\cite{scheele2023fast,ling2022motion,Unhelkar2018HumanAwareRA}. Similar to social navigation, almost all human intent predictors in the context of collaborative manipulation~\cite{Hoffman2023InferringHI, Gui2018TeachingRT,laplaza2021attention,zhang2020recurrent,laplaza2022contextattention, Prasad2022MILDMI, Li2021DirectedAG} generate marginal predictions independent of robot actions. Mainprice et al.~\cite{mainprice2016goal} use motion capture data of two humans engaged in a collaborative task within a shared workspace to predict single-arm reaching motions. However, they do not use any human-robot data to align human and robot representations.

{\bf Human-Robot Correspondence.} 
Our objective is to collect a dataset of paired human-robot interaction. However, it is challenging to program a reliable robot policy that can operate safely with humans. Besides, we require a method to transfer prediction models trained on human-human data to our collected data. When robot and human morphologies match, as in the case of humanoid robots~\cite{valls2024echo, Pollard2002AdaptingHM, Peng2018SFVRL} or dexterous hands \cite{Handa2019DexPilotVT,GarciaHernando2020PhysicsBasedDM}, human motion can be directly imitated. While robotic arm joint movements differ from human joints, a lower-dimensional mapping must be created~\cite{Bharadhwaj2023ZeroShotRM}. In this work, we define a correspondence between the human hand and wrist joints and the robot's end effector to be able to control robots for collecting interaction data. We detect the 6-DoF pose of the human hand using an OptiTrack motion capture suit for robust pose detection amidst occlusions in the environment. Then, we map the pose to the robot's end-effector and control it using an IK-based controller. Similar to the human-robot alignment function used by MimicPlay~\cite{wang2023mimicplay}, an imitation learning framework from human demonstrations, we use the pairing between the robot's motion and the tele-operating human's motion to transfer our prediction model.

\section{Problem Formulation}
\begin{figure*}[t!]
    \centering
    \includegraphics[height = 6cm]{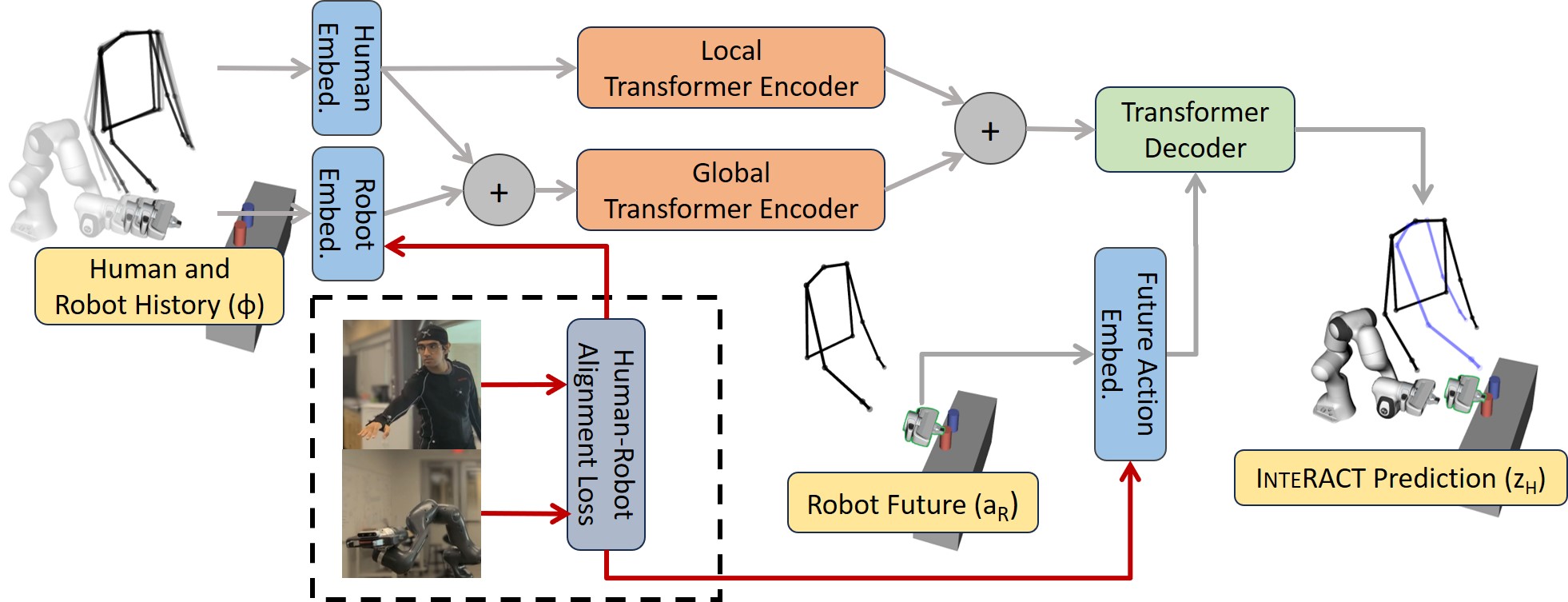}
    \caption{\textbf{\textsc{InteRACT} Model Architecture}. The scene history $\phi$ is encoded by the local and global transformer encoders. The future action $a_{R}$ of the robot is passed as a query to the transformer decoder to generate an \textbf{action-conditioned} human intent prediction $z_{H}$. The robot pose embeddings are aligned with paired human pose embeddings via an \textbf{alignment loss}.}
    \label{fig:model_arch}
    \vspace{-4mm}
\end{figure*} 

\textbf{Marginal Intent Prediction.}
We begin by modeling the marginal intent prediction problem as predicting the human's intent given the scene context. For simplicity\footnote{Our framework can easily be extended to handle multiple humans, a richer context that includes environment information and visual cues, and alternate intent definitions that are a function of the observations.}, we assume there is a single human $H$ interacting with a robot $R$. 

We define the human's \textbf{\emph{intent}} $z_H$ as a T-horizon sequence of future poses, i.e., $z_H = \{s^H_{1}, s^H_{2}, \dots, s^H_{T}\}$, where $s^H_t \in \mathbb{R}^d$ is the $d$-dimensional human pose at future timestep $t$. For many tasks, this information is sufficient for the robot to plan it's future actions. We define \textbf{\emph{context}} $\phi$ as salient information in the scene to predict the human intent. We set context as the current and past history of human states $\phi = \{s^H_{-T+1}, s^H_{-T+2}, \dots, s^H_{0}\}$. We define \textbf{\emph{marginal intent prediction}} model $P_\theta (z_H | \phi)$ as predicting human intent $z_H$ conditioned only on the context $\phi$, where $\theta$ are parameters of the model. Notably, the predictions are independent of the robot R. We train this model via Maximum Likelihood Estimation (MLE) on observed human motions.

\begin{equation}
    \label{eq:marginal_forecasting}
    \max_\theta  \;\mathbb{E}_{\phi, z_H} \log P_\theta(z_H | \phi)
\end{equation}

\textbf{Action-Conditioned Intent Predictions.}
We hypothesize that marginal model $P_\theta(z_H | \phi)$ is insufficient for accurate intent prediction in close-proximity interactions and requires conditioning on robot actions. We define the robot's \textbf{\emph{action}} $a_R \in \mathbb{R}^j$ to be its planned goal location denoted as the $j-$ dimensional robot goal pose. For instance in Fig.~\ref{fig:fig1}, the current poses (and a 1s history) of both the human and robot represents $\phi$, the robot's planned future end-effector position is denoted by $a_R$, and the future human intent $z_H$ is its future upper body pose. We define an \textbf{\emph{action-conditioned intent prediction}} model $P_\theta (z_H | \phi, a_R)$ as predicting human intent $z_H$ conditioned on both the context $\phi$ and the robot action $a_R$. We augment the context to also include the past history of robot states, $\phi = \{s^H_{-T+1}, \dots, s^H_{0}, s^R_{-T+1}, \dots, s^R_{0}\}$, where $s^R_t \in \mathbb{R}^j$ is robot pose at time $t$. We also train this model via a similar MLE objective:

\begin{equation}
    \label{eq:conditional_forecasting}
    \max_\theta  \;\mathbb{E}_{\phi, z_H} \log P_\theta(z_H | \phi, a_R)
\end{equation}

However, optimizing the objective above poses an important practical challenge - collecting large-scale paired human-robot interaction data is costly. It requires humans to work around robots that are already planning reasonable motions. It is also non-trivial to make use of existing public human-human motion datasets to aid in this task. We address both of these challenges next in our approach.

\section{Approach}

We present \textsc{\textbf{InteRACT}} (\textbf{Inte}nt Prediction via \textbf{R}obot \textbf{A}ction-\textbf{C}onditioned \textbf{T}ransformer), a framework for predicting human intent conditioned on future robot actions for collaborative manipulation. At train time, we first pre-train a conditional intent prediction model on human-human interaction data combining publicly available datasets and task specific datasets that we collect. We then fine-tune this model on a small scale human-robot dataset where we predict human intent conditioned on robot actions. Our approach has two main features: (1) an alignment loss between human and robot representations to allow transfer between domains (2) a new tele-operation technique to control a 7-DoF robot arm for paired human-robot interaction.


\subsection{Data: Collecting Paired Human-Robot Interaction \label{sec:approach:correspondence}} We make use of large-scale single-human activity data (AMASS \cite{AMASS:ICCV:2019}) as well as extend the human-human dataset in CoMaD \cite{kedia2023manicast} as our source of human-human interaction data. In order to transfer our action-conditioned model for collaborative manipulation, we further require a dataset of paired human-robot interactions. However, it is not easy to design a robot policy that can be deployed alongside a human partner. To control a robot arm with natural arm movements, we develop a low-level correspondence between the human and the robot. Specifically, we map the human hand's 3-D position as a translation and use the 3-D rotation from the human wrist joint to the hand joint to generate a 6-D end-effector pose for the robot. We track this end-effector pose using an IK-based joint impedance controller\cite{Zhang2020AMR}. Our tele-operation system utilizes an Optitrack Motion capture system that detects human joint positions at 120Hz and can track the calculated 6-D end-effector pose in real-time. We collect not only the joint positions of the robot and its human partner but also the robot-paired joint positions of the tele-operating human. The paired data allows us to align human and robot representations for effective transfer learning (Section \ref{sec:approach:align}). More details included in Section \ref{sec:experiments}.

\subsection{Model Architecture: Action-Conditioned Transformer}


\textbf{Encoding the Scene Context.} 
Fig \ref{fig:model_arch} gives an overview of \textsc{InteRACT}'s model architecture, which is based on Multi-Range Transformer (MRT)~\cite{Wang2021MultiPerson3M}. Both the human history $\in \mathbb{R}^{T \times d}$ and robot history $\in \mathbb{R}^{T \times j}$ (when training on human-human data, the dimensions of both histories are the same) are passed through linear layers and projected to the same embedding dimension $\in \mathbb{R}^{T \times D}$. The human history is passed through a \emph{local} transformer encoder, whereas the combined human and robot history is passed through a \emph{global} transformer encoder. To form the final scene context encoding, both the local transformer encoding $\in \mathbb{R}^{\times T \times D}$ and the global transformer encoding $\in \mathbb{R}^{2 \times T \times D}$ are concatenated together $\in \mathbb{R}^{3 \times T \times D}$. Note that prior to any values being passed into the encoders, a Discrete Cosine Transform (DCT) is applied to them, and an Inverse Discrete Cosine Transform (IDCT) is applied to the final decoder outputs. This practice was introduced by \cite{mao2019learning} to enforce smoothness and periodicity in generated pose outputs.

\textbf{Decoding Human-Intent using Action-Conditioning.} 
MRT decodes future human intent by passing an embedding of the last observable human pose $\in \mathbb{R}^{1 \times d}$ as a query to a Transformer Decoder. In this work, we offset the entire scene around the last human observable pose (and add this offset back into the final predictions). Instead of the last observable human pose, we pass in the robot's future action $a_R \in \mathbb{R}^{1 \times j}$ embedding as the query. When training on human-human data, the human pose 1s in the future $a_H \in \mathbb{R}^{1 \times d}$ is passed in instead. The future action is passed through a linear layer and projected to the same embedding dimension as the encoded contexts $\in \mathbb{R}^{1 \times D}$. This future action embedding is passed in as the query to the transformer decoder.  The scene context encoding vector forms the key and value for the transformer decoder. The decoder output $\in \mathbb{R}^{1 \times D}$ is first passed through a sequence of linear layers to generate a $T$-horizon embedding $\in \mathbb{R}^{T \times D}$. Finally, a linear layer decodes the embedding vector to the human's joint dimensions $\in \mathbb{R}^{T \times J}$. Note that the only change in our architecture from MRT is the query to the transformer decoder.

\subsection{Aligning Human and Robot Representations \label{sec:approach:align}}

\textbf{Representation Mismatch}.
As mentioned in the previous section, the robot and human have different joint dimensions. Besides, they represent different morphologies. In our transformer model, they are projected into $D$-dimensional embeddings via different linear layers. We wish to align the embeddings from human and robot motion into the same embedding space. For this purpose, we utilize the paired data stored during tele-operation while collecting human-robot data. For each robot pose, $s_R\in R^{j}$, we have a corresponding human body pose $s_H\in R^{d}$. We create a dataset $D_{HR}$ from the paired human and robot poses and use it for aligning human-robot representations. 

\textbf{Alignment Loss}. 
To transfer our model from human-human to human-robot data, the learned human and robot embeddings need to be aligned.  We leverage the dataset $D_{HR}$ of paired human and robot poses for this purpose. Specifically, for our transformer model parameterized by $\theta$, we wish to align the robot history embedding layers, parameterized by $\theta_{hist}^H$ and $\theta_{hist}^R$, where the former is utilized to embed human-history when training on human-human interaction data and the latter is used with human-robot data. Concretely, we employ a simple cosine similarity \cite{aytar2017see} loss for the history embedding vectors as follows:
\begin{equation}
\label{eq:align_loss}
L_{align}^{hist}(\theta_{hist}) = \sum_{s_R, s_H}^{D_{HR}}\left[1-S_C(f_{\theta_{hist}^R}(s_R), f_{\theta_{hist}^H}(s_H))\right] 
\end{equation}
where $S_C$ is the cosine similarity metric between two embedding vectors. Similarly, we also align the future-action embedding layers, parameterized by $\theta_{fut}^H$ and $\theta_{fut}^R$.

\textbf{Overall Loss Equation.}
Our complete loss function is therefore the following: 
\begin{equation}
\label{eq:total_loss}
L(\theta) =  \lambda_{p}L_{pred}(\theta) + \lambda_{h}L_{align}^{hist}(\theta_{hist}) + \lambda_{f}L_{align}^{fut}(\theta_{fut}) 
\end{equation}
where $L_{pred}$ is the prediction loss (MPJPE) on the forecasts
\begin{equation}
\label{eq:pred_loss}
L_{pred}(\theta) = \frac{1}{T}\sum_{t=1}^{T}\left\| \hat{s}_{t}^{H}-  s_{t}^{H} \right\|_2^2
\end{equation}
Here, $\hat{s}_{t}^{H}$, $\hat{s}_{t}^{H}$ are the predicted and ground truth human poses respectively. $\lambda_{p}$, $\lambda_{h}$, and $\lambda_{f}$ are loss coefficients (set to 1, 0.1, and 0.1 respectively). Note that there are two separate alignment loss terms, one to indicate the alignment of history motion and one for the alignment of future poses. 





\section{Experiments \label{sec:experiments}}
\begin{figure}[t!]
    \centering
    \includegraphics[height=8.8cm]{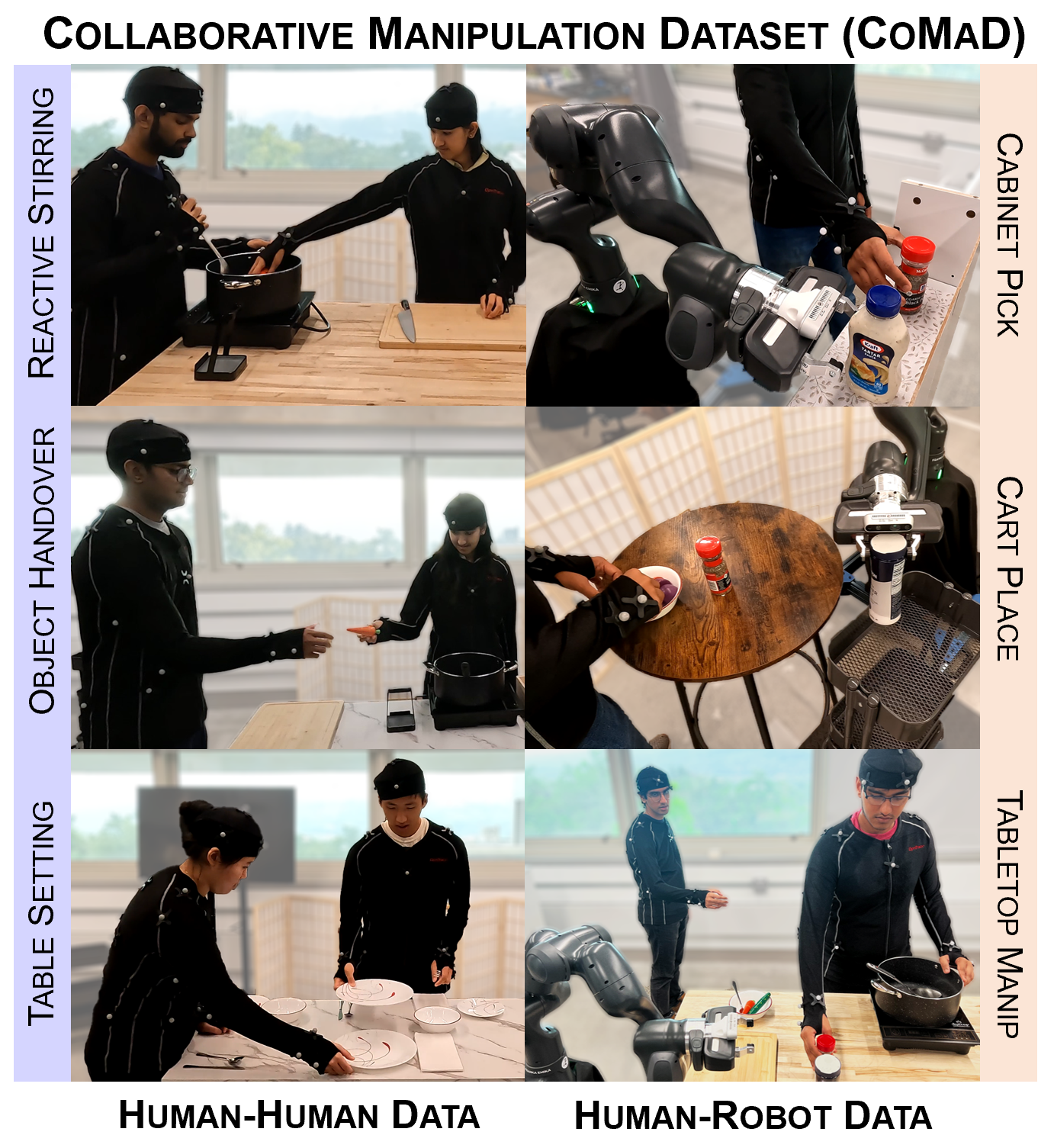}
    \vspace{-2mm}
    \caption{\small \textbf{Collaborative Manipulation Dataset (CoMaD)} consists of Human-Human and Human-Robot interaction data. We collect data on three different H-H tasks and three different H-R tasks across several subjects. The bottom right image shows our tele-operation setup for paired human-robot data collection.}
    \label{fig:dataset}
    \vspace{-6mm}
\end{figure}
\begin{figure*}[t!]
    \centering
    \includegraphics[height=3.5cm]{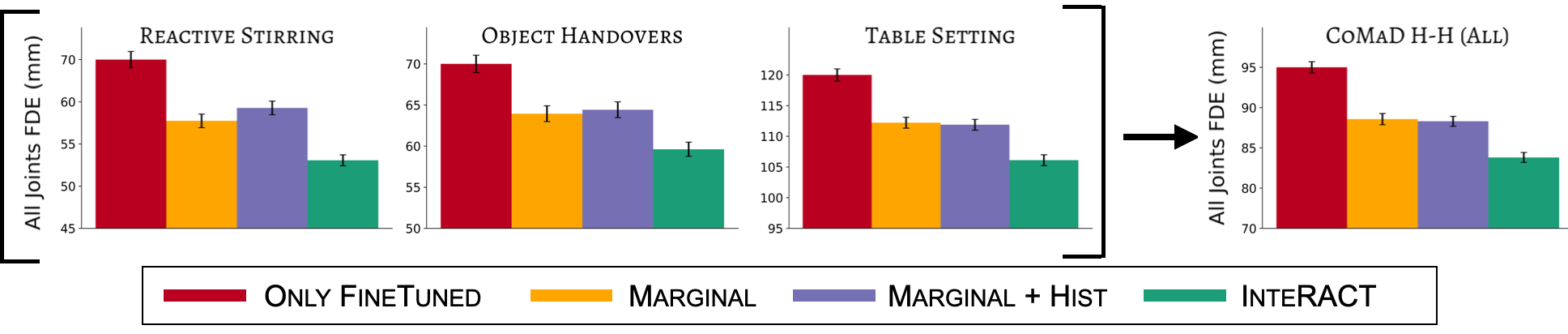}
    \caption{\small All Joints Final Displacement Error (FDE) across all tasks in CoMaD H-H. \textbf{\textsc{InteRACT} predictions have lowest FDE}.}
    \label{fig:h_metrics}
    \vspace{-4mm}
\end{figure*} 



\begin{figure*}[t!]
    \centering
    \includegraphics[width=0.8\textwidth, height=5.45cm]{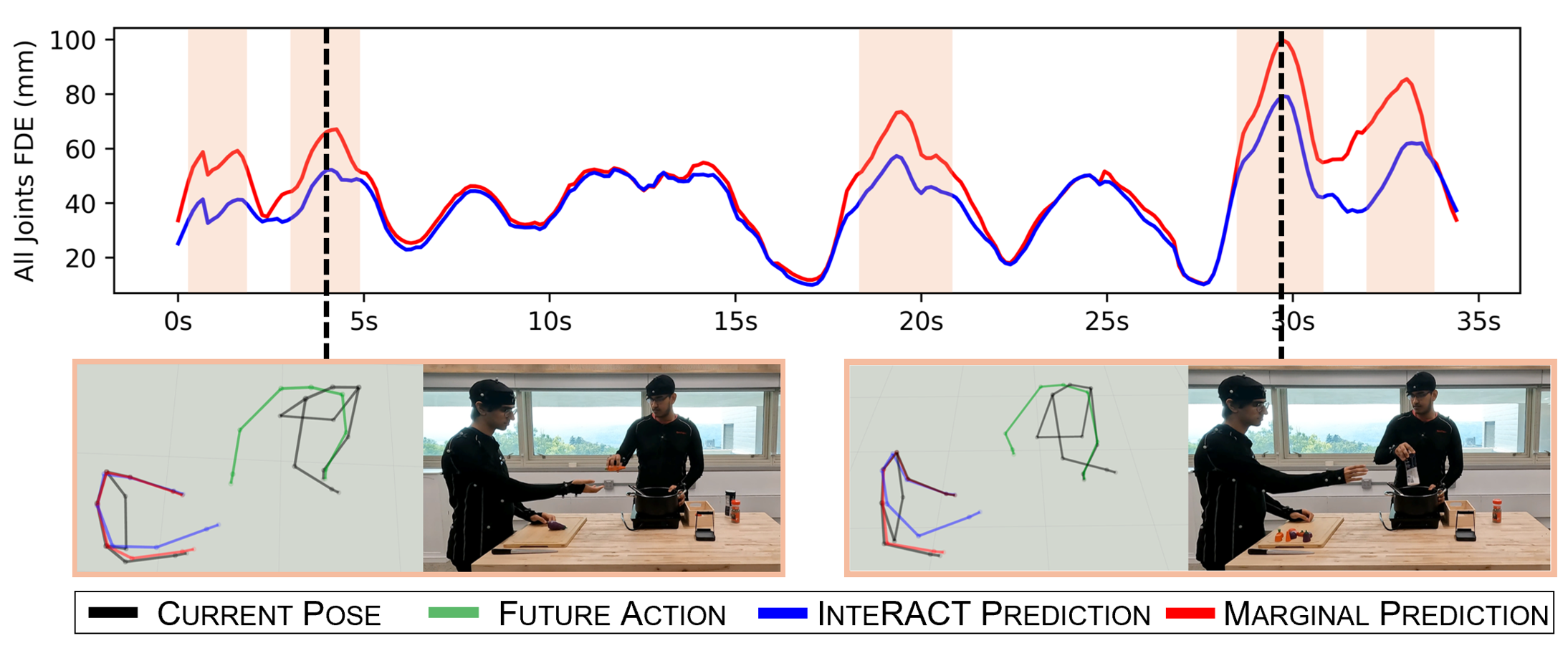}
    \caption{\small \textbf{Top:} Final Displacement Error (FDE) of all joints over time in a test-set episode of object handover. Highlighted windows indicate all object handovers in the episode, where we observe higher errors for \textsc{Marginal}. \textbf{Bottom:} Visualizations of the predictions when the error is at its peak (1s pre-RGB image) show \textsc{InteRACT} anticipates the other's human action and moves towards the handover location.}
    \label{fig:handover_snaps}
    \vspace{-4mm}
\end{figure*} 




\begin{figure*}[t!]
    \centering
    \includegraphics[ height=3.5cm]{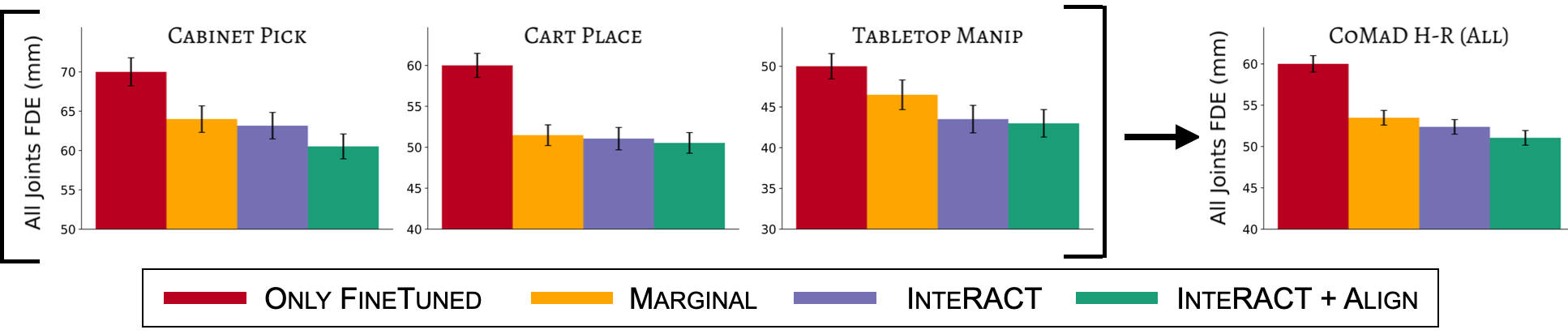}
    \caption{\small Final Displacement Error (FDE) on all joints per and across all tasks in CoMaD H-R. \textsc{InteRACT} variants perform better than other models, with reductions in FDE across tasks with human-robot representation alignment.} 
    \label{fig:hr_metrics}
    \vspace{-4mm}
\end{figure*} 



\begin{figure}[t!]
    \centering
    \includegraphics[width=0.9\columnwidth, height=5.9cm]{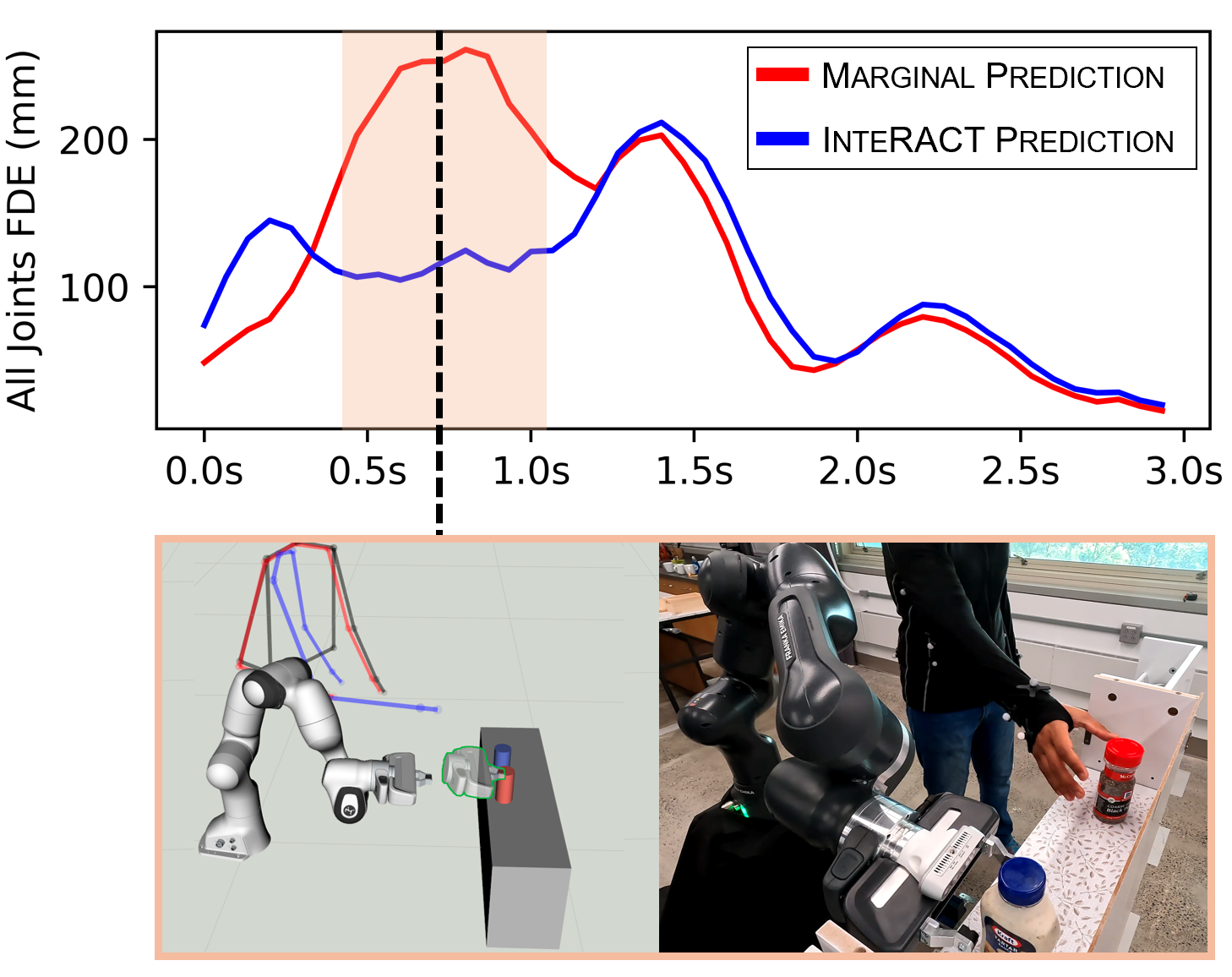}
    \caption{\small Comparing Final Displacement Error (FDE) between \textsc{InterACT} and \textsc{Marginal} predictions in a test-set CoMaD H-R Cabinet Pick episode. \textsc{InteRACT} produces more accurate predictions when the planned robot action is picking up a specific item, indicating the other item is free to pick.}
    \label{fig:hr_cabinet_snaps}
    \vspace{-4mm}
\end{figure} 
\subsection{Collaborative Manipulation Dataset (CoMaD) }
Previous multi-pose prediction methods~\cite{Wang2021MultiPerson3M, vendrow2022somoformer} train and evaluate on small-scale datasets. For example, the human-human interaction split of CMU-Mocap~\cite{cmumocap} consists of just 55 episodes of two human interactions with an average length of 3 seconds, totaling 6 minutes of human motion compared to 40 hours of single-human motion in AMASS. In fact, these methods train their models by augmenting existing datasets with synthetic single-person activity. 

In this paper, we extend the \textbf{Collaborative Manipulation Dataset \textsc{(CoMaD)}}~\cite{kedia2023manicast}. The human-human interaction dataset (Fig \ref{fig:dataset}.) now includes 8 diverse subjects performing 3 different kitchen tasks with a total of 270 episodes (average 30s length), totaling more than 4 hours of human motion. Further, we introduce the human-robot dataset consisting of 217 episodes of interaction collected via tele-operation of a 7-DoF Franka-Emika Research 3 robot with a human partner (Section~\ref{sec:approach:correspondence}). Episodes of each task are divided into train, validation, and test splits in an 8:1:1 ratio.

\textbf{Human-Human Data.} The length of each episode ranges from 20 to 40 seconds. The dataset consists of three tasks: (1) \textsc{Table Setting} (70 episodes): Two humans manipulate table items, avoiding collisions between them. (2) \textsc{Object Handover} (106 episodes): One human asks for objects, and the other human moves in to hand the object over. (3) \textsc{Reactive Stirring} (94 episodes): One human stirs a pot and reacts to the other human pouring vegetables into it.

\textbf{Human-Robot Data.} The length of each episode ranges from 3 to 15 seconds. The dataset consists of three tasks: (1) \textsc{Cabinet Pick} (135 episodes): The robot reaches for one of two objects on the cabinet, and the human responds accordingly. If the robot reaches for the object close to the human, they wait, and if the robot reaches for the object away from the human, both reach for their respective objects. (2) \textsc{Cart Place} (55 episodes): There is a table and cart setup between the robot and the human. The robot moves an object from the table to the cart, and the human picks up an object from the cart to use. If the human moves first, the robot must wait for the human, and if the robot moves first, the human must wait for the robot. (3) \textsc{Tabletop Manipulation} (27 episodes): A table has two objects on it, with both the human and robot reaching for one of them. The human waits for the robot when it moves in. Similarly, the robot must wait if the human comes in the way.

\subsection{Experimental Setup}
\textbf{Large Human-Activity Databases.} We created synthetic two-human data using AMASS~\cite{AMASS:ICCV:2019} and pre-trained the model using the synthetic data and CMU-Mocap~\cite{cmumocap} data. We use the human-human interaction data in CMU-Mocap without adding any synthetic humans.


\textbf{Baselines (H-H).} \textsc{Marginal}~\cite{kedia2023manicast} uses one human's history to predict intent, whereas \textsc{Marginal (+ Hist)}~\cite{Wang2021MultiPerson3M} also uses the other human's history. Both are pre-trained on synthetic AMASS data and fine-tuned on H-H data. \textsc{Only FineTuned} is only trained on a smaller amount of H-H data. Our method, \textsc{InteRACT} uses both humans' histories and conditions on the other human's future action.

\textbf{Baselines (H-R).} {\textsc{Marginal}} takes the corresponding H-H model above and fine-tunes on H-R data, whereas \textsc{Only FineTuned} is only trained on H-R data. {\textsc{InteRACT}} takes our H-H model and fine-tunes on H-R data, replacing the second human's encoding with the robot. {\textsc{InteRACT + Align}} further incorporates the robot alignment loss (Eq \ref{eq:align_loss}). 




\textbf{Implementational Details.} 
We utilize a 1s motion history input to generate a 1s forecast (represented over 15 timesteps). We consider the human pose dimension $d=27$, which includes 9 upper body 3-D joint positions (upper back, shoulders, elbows, wrists, hands), and the robot pose dimension $j=6$, which includes two 3-D points on the robot's end-effector corresponding to the human's hand and wrist. We report the Final Displacement Error (FDE), which is the average distance between the predicted joint positions and ground truth joint positions at the end of 1s.

\subsection{Results and Analysis}
\textbf{\textit{O1.} Conditioning on actions improves intent prediction in both human-human and human-robot interactions.} Fig.\ref{fig:h_metrics} and Fig.\ref{fig:hr_metrics} both show that \textsc{InteRACT} models outperform any \textsc{Marginal} models without information about the intent of the other agent in the scene. \textsc{Marginal} models produce higher FDE on all three H-H and H-R tasks compared to conditional models. This can be seen qualitatively in Fig.\ref{fig:handover_snaps} where the \textsc{InteRACT} intent predictions anticipate a handover due to knowledge about the planned action of the other human in the scene. Similar trends follow in H-R tasks such as \textsc{Cabinet Pick} demonstrated in Fig \ref{fig:hr_cabinet_snaps} where conflict arises as a human and robot simultaneously reach for objects. If the robot reaches for the object on the right, we know the human intends to pick the object on the left.

\textbf{\textit{O2.} Human-Robot Alignment loss helps improve prediction performance.} Fig.\ref{fig:hr_metrics} shows that adding alignment loss (\textsc{InteRACT + Align}) reduces FDE in predicting future human poses. This supports our hypothesis that aligning representations helps in transfer learning from H-H data.

\textbf{\textit{O3.} Pre-training models on human-human interactions is critical for transfer learning.} Fig.\ref{fig:hr_metrics} shows that \textsc{Only FineTuned} trained only on H-R data performs significantly worse than other \textsc{Marginal} and \textsc{InteRACT} that are also trained on H-H data. It yields notably higher FDE across all joints in all three H-R tasks we evaluate on.  

\textbf{\textit{O4.} Pre-training on synthetic human-human activity data helps learn general human motion dynamics.} Fig.\ref{fig:h_metrics} shows that \textsc{Only FineTuned} produces higher FDE than models pre-trained on synthetic AMASS data despite the synthetic data lacking real H-H interactions. This leads us to believe that large-scale single-human data can be leveraged even in the multi-human setting.

\section{Discussion and Limitations}
In this work, we present \textsc{InteRACT}, a novel architecture that predicts human intentions by \textbf{conditioning on future robot actions}. We also expand the Collaborative Manipulation Dataset (CoMaD) with a novel \textbf{paired human-robot dataset} collected by tele-operation allowing us to effectively \textbf{align} a model trained on human-human data to human-robot interactions. In the future, we aim to demonstrate the performance of \textsc{InteRACT} in online planning scenarios. By reasoning about how actions can influence human intent, robots can be more confident in their plans.

\textbf{Limitations.} There are notable limitations to our work that we highlight in this section. Robot safety in close proximity interactions is extremely important, and collisions can be a concern in the case of errors in human intent prediction. Safety mechanisms \cite{lasota2017survey} studied extensively should be used to help target these potential issues. While we collect data across several subjects, we are limited to certain environments per task. Our goal is to collect data in a distribution that represents a few different modes of motion that are common in human-robot interactions, and plan to expand the dataset in the future to cover a wider distribution.

\section{Acknowledgements}
This work was partially funded by NSF RI (\#2312956).

\bibliographystyle{IEEEtran}
\bibliography{IEEEabrv,refs}

\end{document}